\documentclass{bmvc2k}

\title{Do Saliency Models Detect Odd-One-Out Targets? New Datasets and Evaluations}

\addauthor{Iuliia Kotseruba}{yulia_k@eecs.yorku.ca}{1}
\addauthor{Calden Wloka}{cwloka@eecs.yorku.ca}{1}
\addauthor{Amir Rasouli}{aras@eecs.yorku.ca}{1}
\addauthor{John K. Tsotsos}{tsotsos@eecs.yorku.ca}{1}

\addinstitution{
 Department of Electrical Engineering and Computer Science\\
 York University\\
 Toronto, Canada
}
\runninghead{Kotseruba, \etal}{Do Saliency Models Detect Odd-One-Out Targets?}

\usepackage{multirow}
\usepackage{subfigure}
\usepackage{hyphenat}
\def\eg{\emph{e.g}\bmvaOneDot}

\def\etal{\emph{et al}\bmvaOneDot}

\newcommand{\degree }{$^{\circ}$}
\newcommand{\3}{$^3$}

\begin{document}

\maketitle

\begin{abstract}
Recent advances in the field of saliency have concentrated on fixation prediction, with benchmarks reaching saturation. However, there is an extensive body of works in psychology and neuroscience that describe aspects of human visual attention that might not be adequately captured by current approaches. Here, we investigate singleton detection, which can be thought of as a canonical example of salience. We introduce two novel datasets, one with psychophysical patterns and one with natural odd-one-out stimuli. Using these datasets we demonstrate through extensive experimentation that nearly all saliency algorithms do not adequately respond to singleton targets in synthetic and natural images. Furthermore, we investigate the effect of training state-of-the-art CNN-based saliency models on these types of stimuli and conclude that the additional training data does not lead to a significant improvement of their ability to find odd-one-out targets.

\end{abstract}

\section{Introduction}
\label{sec:intro}
The human visual system processes vast amounts of incoming  information and performs multiple complex visual tasks with perceived efficacy and ease. However, it is well known that not only are humans capacity-limited perceivers \cite{Broadbent1958}, but also that the vision problem in general is computationally intractable \cite{tsotsos_1990}. Instead of inspecting every element of a scene, humans use a set of attention mechanisms to prioritize and filter stimuli from the early visual processing stages through to the higher visual areas. The ability to use perceived saliency of objects for efficient scanning of the scene is one of the fundamental attention mechanisms.

Computational approaches to predicting human judgements of saliency have resulted in an extensive corpus of saliency models. The evaluation for these models in recent years has been conducted using a number of metrics for fixation prediction \cite{mit-saliency-benchmark}, however, such heavy focus on gaze prediction fails to address extensive research in psychology and neuroscience which has catalogued and quantified many aspects of human visual attention \cite{bruce2015computational}.

One area well studied in psychology is visual search, frequently associated with Feature Integration Theory (FIT) \cite{treisman1980feature}. FIT posits that items that differ significantly in one feature dimension are processed in a parallel manner, whereas an item uniquely defined by a combination of features requires a serial search through items. Wolfe \cite{Wolfe1998,wolfe2017five}, on the other hand, notes that the parallel/serial dichotomy of FIT is better represented as a continuum of search efficiencies. In other words, if the difference between the target and surrounding distractors is sufficiently large, the target will be immediately detected (``pop-out''), otherwise as the target-distractor difference decreases search becomes increasingly inefficient \cite{wolfe1998can,wloka2017interaction}.

Visual search was a major guide in the early formulation of saliency models (\eg \cite{KochUllman1985, itti1998model, Rosenholtz1999}). Thus, given its conceptual importance to the topic of saliency, here we propose two new datasets with synthetic and natural odd-one-out targets and a large-scale comparative study that systematically evaluates the responses of classical and deep-learning saliency models to these types of stimuli. These results not only identify and quantify remaining gaps between human and algorithmic performance in visual search, but also have significant practical ramifications. Currently, saliency models are being applied in commercial areas, such as the design of marketing materials (\eg \cite{PietersEtAl2007,vanderLansEtAl2008,RosenholtzEtAl2011, VAS_Sample2015,WastlundEtAl2018}),
where strong discrimination of specific information or products is a clearly stated goal. Thus the inadequate performance of saliency models on a range of realistic odd-one-out targets revealed by our evaluations highlights urgent new research avenues for these application areas.

\subsection{Relevant Works}
\label{sec:relevant_works}
\textbf{Psychophysical Evaluation of Saliency Models.}
Although a number of saliency models present qualitative results on a small set of synthetic stimuli, such as color/orientation singletons, Q/O and visual search asymmetries \cite{bruce2007information, hou2007saliency, zhang2008sun,  seo2009static, goferman2012context, Erdem2013, hou2012image, Riche2013, Kummerer2014, vig2014large, huang2015salicon, bruce2016deeper, kummerer2017understanding}, only a few have been systematically evaluated on the psychophysical patterns used in human experiments. For instance, IKN has been evaluated on color/orientation pop-out search and conjunctive search  \cite{itti2000saliency}, AWS and Discriminant Saliency were tested on non-linearity of pop-out for orientation and various search asymmetries \cite{garcia2012saliency, gao2008plausibility}.

Bruce et al. \cite{bruce2015computational} discussed a range of psyhological factors influencing visual saliency and their modeling implications. Additionally, several studies systematically compared the performance of multiple saliency algorithms against one another and human gaze data. Borji et al. \cite{borji2013quantitative} conducted a quantitative analysis of 35 classical saliency models over 54 synthetic patterns from the CAT2000 dataset \cite{CAT2000} and compared them to human fixation data. Wloka et al. \cite{wloka2016psychophysical} compared human response time data against a set of classical saliency models on singleton search with oriented bars. More recently, Berga et al. \cite{berga2018neurodynamical} evaluated a large set of models (both classical and deep) on CAT2000/Pattern image set and synthetic patterns \cite{berga2019psychophysical}.

Together these studies suggest that no existing saliency model accounts for most of the identified saliency phenomena, although there is evidence that models based on the FIT framework \cite{treisman1980feature} may perform better on synthetic images \cite{borji2013quantitative}. While pointing to the same conclusion, these studies are difficult to compare directly as they all use different metrics, sets of algorithms and datasets for evaluation.

\textbf{Psychophysical Patterns and Odd-One-Out Images in Datasets.}
Datasets typically used in saliency research include only a handful of synthetic psychophysical patterns, if any. For instance, the Toronto dataset \cite{bruce2007attention} includes 16 such cases out of 136 images,  MIT1003 \cite{judd2009learning} has 4  patterns out of 1003 and CAT2000/Patterns image set \cite{CAT2000} has 71 synthetic patterns out of 2000. The recently proposed SID4VAM  dataset \cite{berga2019psychophysical} provides human fixation data for 230 synthetic patterns covering 15 different stimuli types. However, since human participants were given task instructions, this data is not directly comparable with the free-viewing conditions under which saliency models are more typically tested.

Odd-one-out images, which can be thought of as realistic counterparts to psychophysical patterns (see Figure \ref{fig:oddoneout_samples}), are also rarely found in fixation datasets. Out of 9 datasets that we examined none contained a significant portion of these types of stimuli, \eg only 3 such images in the MIT1003 dataset (1003 images), 19 in DUT-OMRON (5168 images) \cite{yang2013saliency} and 33 in SALICON (20K images)\cite{jiang2015salicon} (see Section \ref{sec:data_stats} for a complete list).

\textbf{Contributions.}  We address the research challenges described above by providing the four following contributions: 1) a large dataset of synthetic search arrays and a set of tools for analysis and evaluation of saliency models; 2) a large annotated dataset of images with odd-one-out objects captured ``in the wild''; 3) extensive analysis of saliency algorithms with respect to several features known to guide human attention; 4) experiments showing the impact of augmenting training data with synthetic and/or real images on the performance of the CNN-based models on odd-one-out task and fixation prediction.

\section{Psychophysical Patterns (P$^{3}$) Dataset}
\begin{figure*}
\centering
\includegraphics[width=0.95\textwidth]{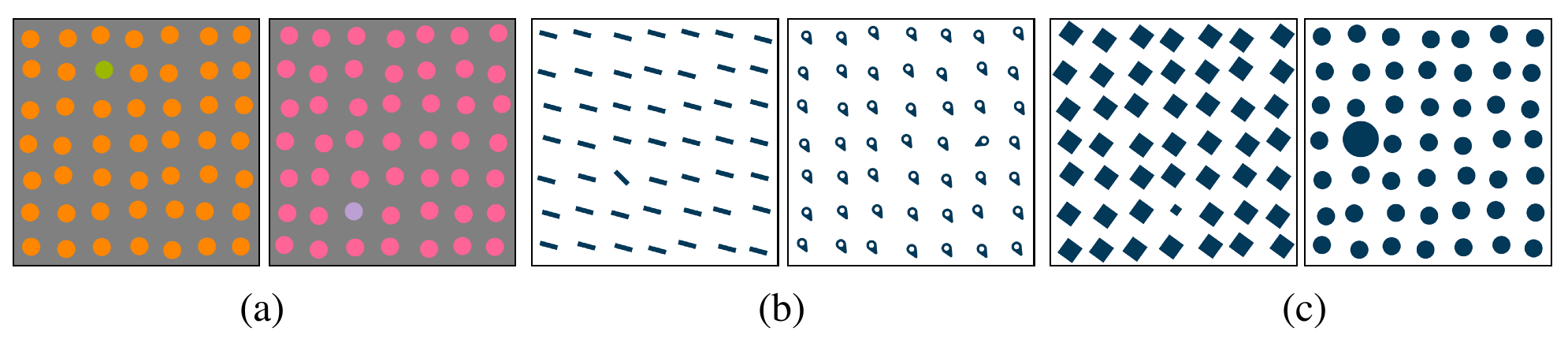}
\caption{Sample images from P\3 with (a) color, (b) orientation and (c) size singletons.\label{fig:psychosal_samples}}
\end{figure*}

We propose a large dataset of psychophysical patterns (P\3)\footnote{\label{note1}\url{http://data.nvision2.eecs.yorku.ca/P3O3/}} for evaluating the ability of saliency algorithms to find singleton targets. We focus on color, orientation, and size: three features recognized as undoubted attention guiding attributes \cite{wolfe2017five} (see Figure \ref{fig:psychosal_samples})

All visual search arrays in the P\3 dataset are on a regular $7\times 7$ grid with distractor size 75 px ($\approx$2\degree) and target size between 18 px (0.5\degree) and 140 px (4\degree). The target location in the grid is selected at random. Jitter of 15 px is applied to each element (as recommended in \cite{arun2012turning} to avoid perceptual grouping effects). The size of images ($1024\times 1024$) matches the setup of MIT300 \cite{mit-saliency-benchmark} ($\approx$35 px per degree of visual angle and image sizes of 768$\times$1024). Each image has a corresponding ground truth mask for the target and distractors. 

There are a total of 2589 images in P\3 including 885 color, 864 orientation and 840 size search arrays. Further details on how the images are generated can be found in Section \ref{sec:data_gen}.

\section{Odd-One-Out (O\3) Dataset}

\begin{figure*}
\centering
\includegraphics[width=0.95\textwidth]{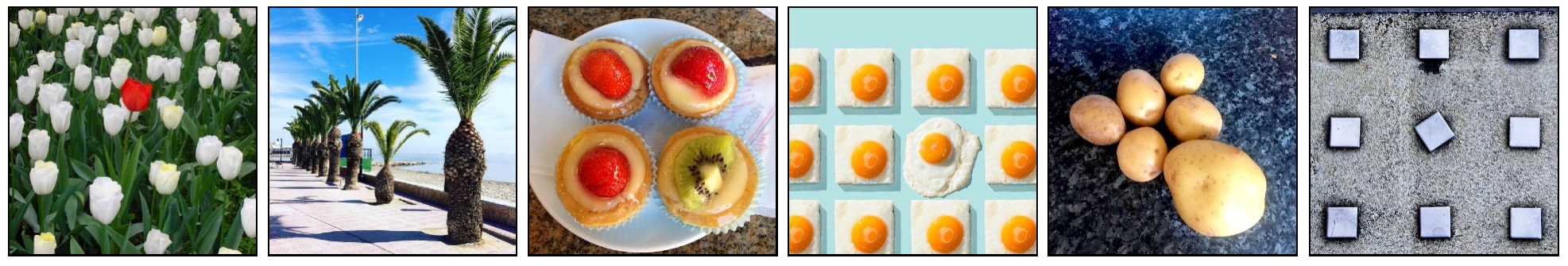}
   \caption{Sample images from the O\3 dataset with singletons in various feature dimensions. From left to right: color, size, color/texture, shape, size, orientation.}
\label{fig:oddoneout_samples}
\end{figure*}

In addition to the psychophysical arrays, we propose a dataset of realistic odd-one-out stimuli gathered ``in the wild''. Each image in the novel Odd-One-Out (O\3) dataset\textsuperscript{\ref{note1}} depicts a scene with multiple objects similar to each other in appearance (distractors) and a singleton (target) which usually belongs to the same general object category as distractors but stands out with respect to one or more feature dimensions (e.g. color, shape, size) according to the evaluation of three human annotators (see samples in Figure \ref{fig:oddoneout_samples}).

The O\3 dataset consists of 2001 images with the larger image dimension set to 1024 to match the setup of MIT300, as most saliency models used for evaluation are optimized for it. Each image is annotated with segmentation masks for targets and distractors as well as text labels for the following: type of the target object, number of distractors and pop-out features (color, pattern/texture, shape, size, orientation, focus and location).

Targets in the O\3 dataset represent nearly 400 common object types such as \textit{flowers}, \textit{sweets}, \textit{chicken eggs}, \textit{leaves}, \textit{tiles} and \textit{birds}. Due to the importance of color as an attention guiding feature \cite{wolfe2010visual} the dataset contains many color singletons. Specifically, 37\% of all targets differ from the distractors by color alone and 47\% differ by color and one or more additional feature dimension. In addition, 33\% of the targets are distinct in texture, 26\% in shape, 19\% in size and 8\% in orientation. The targets vary considerably in size: approximately 80\% of the targets occupy between 7\% and 20\% of the image by area, 30\% of the targets are larger and few take up to 80\% of the image. All targets are accompanied by at least 2 distractors, with half of the targets surrounded by at least 10 distractors and 10\% by over 50 distractors.

\section{Experiments}
\subsection{Metrics}
Reaction time (RT) is a primary metric of human performance reported in the psychophysical literature (\eg \cite{wolfe1994guided,Wolfe1998,arun2012turning}). Although RT cannot be directly measured from saliency maps, there is a link between reaction time and the relative salience of targets and distractors \cite{ToellnerEtAl2011,ZehetleitnerEtAl2013}. We therefore use several proxy measures: the number of fixations before reaching the target, global saliency index (GSI), and the saliency ratio (SR).

\textit{Number of fixations.} To produce fixations from a saliency map we iterate through maxima and suppress attended locations with a circular mask (as in \cite{Wloka_2018_CVPR}) until a location near the target is reached or a maximum number of fixations is exceeded. The circular mask has a 1\degree \ radius which corresponds to the size of distractors in all images.

There is no definition of how close to the target the fixation should land in order to be considered as a hit. Values reported in the literature vary from 1\degree  \ \cite{becker2008mechanism, becker2013simply, becker2010role} to 2\degree \ \cite{tseng2014modeling} or up to 2.35\degree \ \cite{becker2013higher} radius around the center of the target. In our experiments we set the radius to 1\degree \ around the center of the target and up to 2\degree \ for larger stimuli in \textit{size} patterns.

\textit{Global Saliency Index}. The metric is defined as $GSI = \frac{S_{target}-S_{distr}}{S_{target}+S_{distr}}$, where $S_{target}$ and $S_{distr}$ are the average saliency values within the target and distractor masks respectively \cite{spratling2012predictive, soltani2010visual}. GSI measures how well the target is distinguished from the distractors and varies between -1 (target saliency is 0, only distractors are highlighted) to 1 (only the target is highlighted).

\textit{Saliency Ratio}. Due to the highly varying target-distractor sizes and configurations in real pop-out images in the O\3 dataset, both the GSI and fixations-to-target metrics become challenging to apply. We, therefore, use a more straightforward ratio of maximum saliency of the target vs maximum saliency of the distractors as in \cite{wloka2016psychophysical} and the same for the background vs target, referred to in the text as \textit{MSR$_{targ}$} and \textit{MSR$_{bg}$} respectively. Since most practical applications require locations of maxima within saliency maps, these two metrics help determine whether the target is more salient than the surrounding distractors and background.

\subsection{Saliency Models and Datasets}
For all experiments that follow we use 12 classical saliency models, i.e. theory-driven models with clear semantic interpretation (AIM \cite{bruce2006saliency}, BMS \cite{zhang2013saliency}, CVS \cite{Erdem2013}, DVA \cite{hou2009dynamic}, FES \cite{tavakoli2011fast}, GBVS \cite{harel2007graph}, IKN \cite{itti1998model}, IMSIG \cite{hou2012image}, LDS \cite{fang2017learning}, RARE2012 \cite{Riche2013}, SSR \cite{seo2009static} and SUN \cite{zhang2008sun}), and 8 deep learning models, which are primarily data-driven (DGII \cite{kummerer2016deepgaze}, DVAP \cite{wang2018deep}, eDN \cite{vig2014large}, ICF \cite{kummerer2017understanding}, MLNet \cite{cornia2016deep}, oSALICON \cite{huang2015salicon, thomas2016opensalicon}, SalGAN \cite{pan2017salgan} and SAM-ResNet \cite{cornia2018predicting}). 

We use the SMILER framework \cite{wloka2018smiler} to run all models without center bias on P\3 and with center bias on O\3. Default values for other parameters (including smoothing) were kept. We evaluate all models on P\3, O\3 and CAT2000/Pattern.

\subsection{Evaluation of Saliency Models}
\subsubsection{Psychophysical Evaluation of Saliency Models}

Evaluating models on psychophysical search arrays allows us to measure their response to basic attention guiding features and determine whether they capture known properties of the human visual system.

\begin{figure*}
\centering
\subfigure[\% of detected targets vs number of fixations]{
\includegraphics[width=0.48\textwidth]{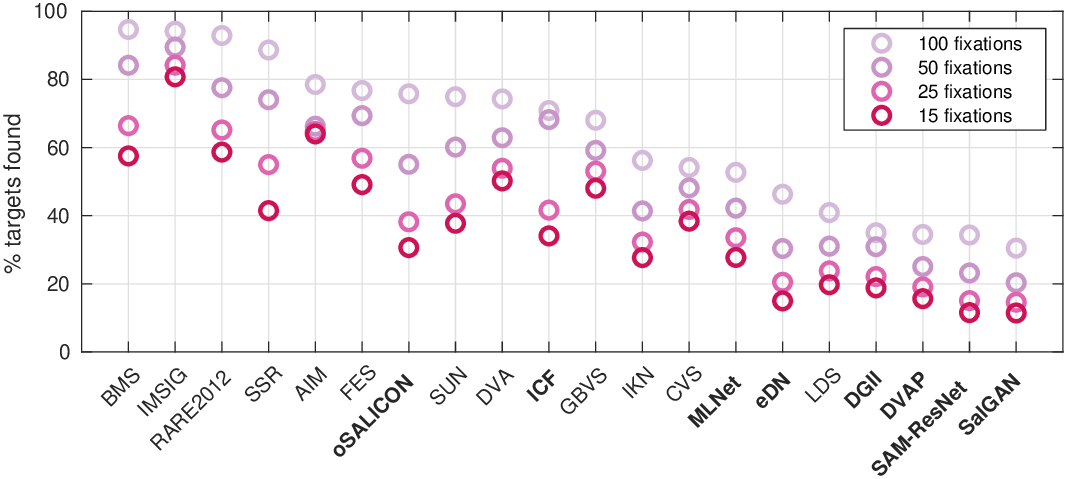}
\label{fig:found_vs_fixations}
}
\subfigure[\% of detected targets vs singleton feature]{
\includegraphics[width=0.48\textwidth]{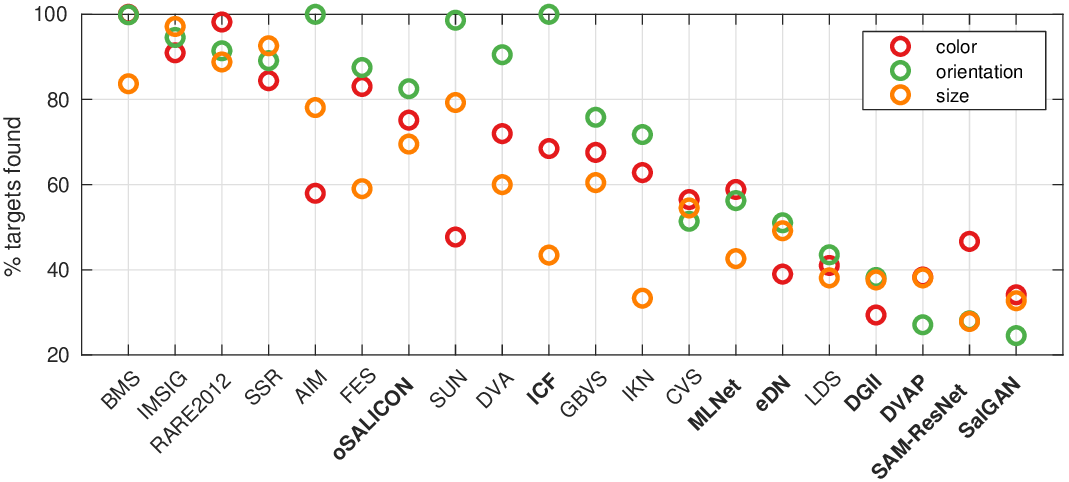}
\label{fig:found_vs_stimuli_type}
}
   \caption{a) Number of fixations vs \% of targets detected.  b) Performance on color, orientation and size singletons at maximum of 100 fixations. Models are sorted by the \% of targets detected at maximum of 100 fixations. Labels for deep models are shown in bold.}
\end{figure*}

\textbf{Target detection rate.} To measure the strength of attention guidance we count the total number of targets detected and the average number of fixations required to do so. The relationship between these quantities is shown Figure \ref{fig:found_vs_fixations}. When the maximum number of fixations is set to 100 (or $2\times$ the number of elements in the grid), several classical algorithms (BMS, IMSIG, and RARE2012) achieve success rate above 90\% and only one deep-learning model (oSALICON) comes close to the 75\% mark. Note that the performance of nearly all models degrades quickly when the allowed number of fixations is reduced.

The same evaluation applied to human fixation maps for 51 singleton patterns from the CAT2000/Pattern yields 98\% target detection rate and 13 fixations on average. Note that the CAT2000 data is not directly comparable with P\3 in image size and stimuli used. However, even when the maximum number of fixations is set to 25 (equivalent to examining half of the elements in the grid), most saliency algorithms miss  $>50\%$ of the targets in the P\3 dataset and only one model, IMSIG, detects slightly more than 80\% of singletons.

\begin{figure*}[t]
\centering

\subfigure[color difference]{
\includegraphics[width=0.31\textwidth]{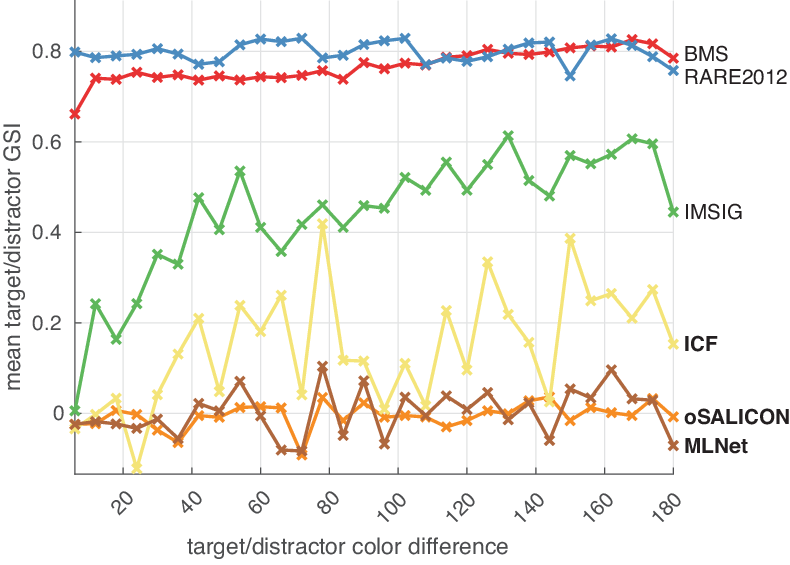}
\label{fig:abs_hue_angle_diff}
}
\subfigure[orientation difference]{
\includegraphics[width=0.31\textwidth]{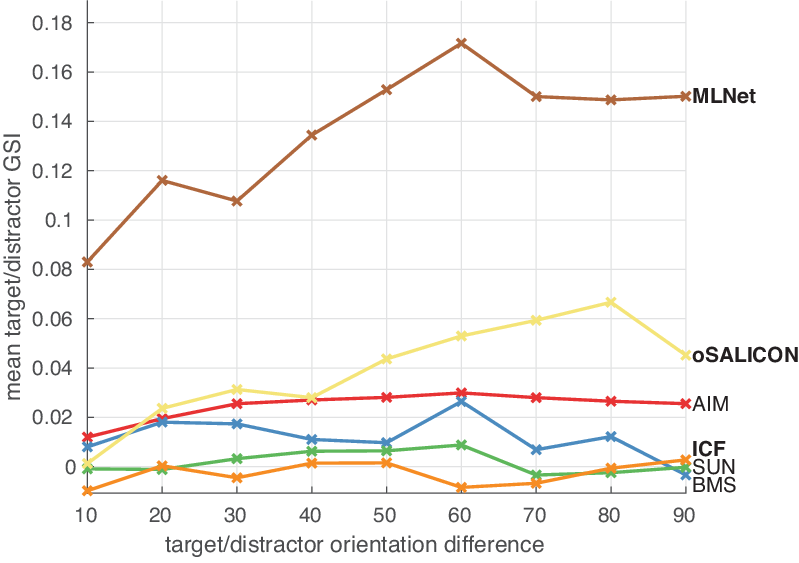}
\label{fig:abs_orientation_diff}

}
\subfigure[size ratio]{
\includegraphics[width=0.31\textwidth]{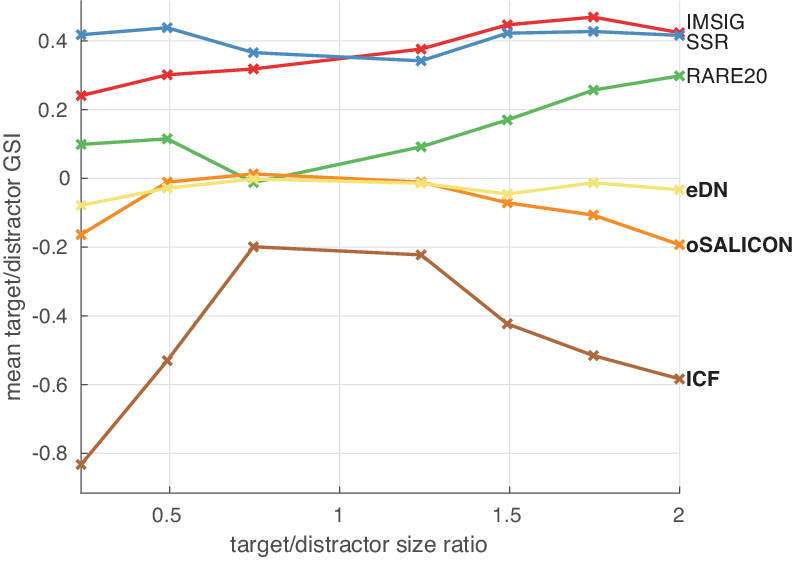}
\label{fig:size_ratio}
}



\subfigure[saliency maps]{
\includegraphics[width=1\textwidth]{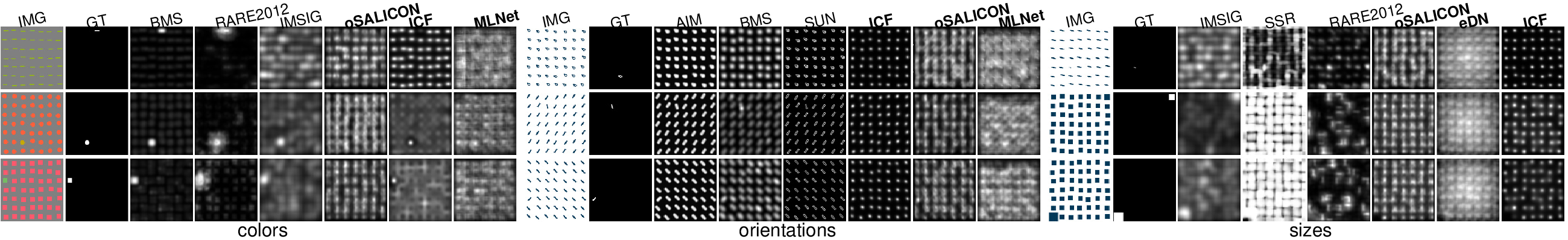}
\label{fig:P3_samples}
}
   \caption{The discriminative ability (GSI score) of the top-3 classical and deep models for a range of TD differences in color (a), orientation (b) and size (c) feature dimensions. The models are selected based on the total number of targets found within 100 fixations in each dimension. d) Sample saliency maps for each model. TD difference for color and orientation targets increases from top to bottom row. Size targets range from the smallest to largest.}
\label{fig:search_feature}
\end{figure*}

\textbf{Response to simple features.} As shown in Figure \ref{fig:found_vs_stimuli_type}, the performance of most models varies significantly depending on the type of stimuli they are applied to. There are, however, several exceptions, namely IMSIG, SSR, CVS and LDS, which perform equally well or equally poorly on all three types of singletons. In general, orientation and color targets appear to be easier to detect for many of the models and especially for those models that process these features explicitly, \eg RARE2012, BMS, AIM. Size targets are missed more often, even by the models that do multi-scale processing (\eg RARE2012, IKN, MLNet).

\textbf{Sensitivity to target-distractor similarity.} It has been established that visual search efficiency improves as the target-distractor (TD) difference increases \cite{wolfe2004attributes}. Above a certain value, the pop-out effect can be observed, i.e. the target is found in nearly constant time (or only few fixations) regardless of the number of the distractors.

The plots in Figure \ref{fig:search_feature} visualize the discriminative ability of saliency algorithms for a range of TD differences in color, orientation and size. Due to the space limitations, we only show and discuss the 3 best classical and deep models for each feature type. Top models are selected based on the \% of the detected targets of each feature type. The plots with GSI measures and fixation counts for all models can be found in Section \ref{sec:psych_eval}.

\textbf{Color singletons.} Among the classical models, only IMSIG has a resemblance to human behavior on color stimuli (Figure \ref{fig:abs_hue_angle_diff}). It has more difficulty distinguishing similar colors and gradually improves as the TD difference increases. However, the behavior of IMSIG is not consistent as indicated by the dips in the GSI plot. Two other classical algorithms, BMS and RARE2012, nearly perfectly find even the smallest differences in hue and detect targets in fewer than 10 fixations on average. Note also very clear discrimination of the targets in the saliency maps (see  Figure \ref{fig:P3_samples}). ICF, the only CNN-based model that explicitly includes color contrast features, shows a tendency of improving on larger TD differences but is less consistent (note the spikes around 80\degree \ and 150\degree). oSALICON and MLNet both have low GSI scores for all TD differences and need more than 50 fixations on average to find even the most salient targets.

\textbf{Orientation singletons.} The GSI scores for orientation targets are nearly an order of magnitude lower than scores for color targets as shown in Figure \ref{fig:abs_orientation_diff}. Here, the top-3 classical models, AIM, SUN and BMS, do not discriminate the targets well. Only AIM has a GSI score consistently above 0 and takes only a few fixations on average to reach the target. SUN and BMS have less consistent GSI scores and perfom $\approx$ 30 fixations to find targets regardless of TD difference. Deep models MLNet and oSALICON detect more distinct targets better (with spikes around 60\degree \ and 70\degree) but require at least $40$ fixations on average to reach the target.

\textbf{Size singletons.} Similar to color and orientation features, we expect that the targets closest to the size of the distractors would be more difficult to find, while very small and very large targets should `pop' easily. Unlike orientation and color, most classical algorithms do not encode size explicitly as a feature but may handle it indirectly via multi-scale processing. Here, IMSIG, SSR and RARE2012 exhibit anticipated behavior to some extent, having higher GSI for larger TD differences (Figure \ref{fig:size_ratio}). Interestingly, deep learning models eDN and oSALICON demonstrate almost the opposite of the human behavior as their GSI scores drop for larger targets. Another deep model, ICF, has negative GSI scores for the whole range of TD differences, meaning that the target is consistently less salient than the distractors on average, as can be seen in the sample saliency maps shown in Figure \ref{fig:P3_samples}.

Based on the evidence presented we conclude that the majority of saliency models are not guided by attributes such as color, orientation and size which have an undoubted impact
on human attention \cite{wolfe2017five}. Several classical algorithms (\eg IMSIG, RARE2012 and AIM) are guided by some of the basic features, but none of them respond to all three features considered. In comparison to classical models, deep algorithms have lower discriminative ability and are less consistent in finding singletons.

\subsubsection{Evaluation of Saliency Models on Odd-One-Out Data}

Unlike the psychophysical evaluation that focused on how well the saliency models capture fundamental properties of the human visual system, testing models on the O\3 data aims to examine their behavior in more realistic scenarios. To the best of our knowledge, ours is the first work that addresses detection of singletons by saliency algorithms in realistic stimuli.

\begin{table}[t]
\centering
\resizebox{0.85\textwidth}{!}{%
\begin{tabular}{l|ll|ll|ll}
\multicolumn{1}{c|}{\multirow{2}{*}{Model}} & \multicolumn{2}{c|}{Color targets}                                   & \multicolumn{2}{c}{Non-color targets}            & \multicolumn{2}{c}{All targets}                                    \\ \cline{2-7}
\multicolumn{1}{c|}{}                       & \multicolumn{1}{c}{MSR$_{targ}$} & \multicolumn{1}{c|}{MSR$_{bg}$} & \multicolumn{1}{c}{MSR$_{targ}$} & MSR$_{bg}$ & \multicolumn{1}{c}{MSR$_{targ}$} & \multicolumn{1}{c}{MSR$_{bg}$} \\ \hline
\textbf{SAM-ResNet}                         & 1.47                          & 1.46                         & 1.04                          & 1.84    & 1.40                          & 1.52                        \\
CVS                                         & 1.43                          & 2.43                         & 0.91                          & 4.26    & 1.34                          & 2.72                        \\
\textbf{DGII}                               & 1.32                          & 1.55                         & 0.94                          & 1.95    & 1.26                          & 1.62                        \\
FES                                         & 1.34                          & 2.53                         & 0.81                          & 5.93    & 1.26                          & 3.08                        \\
\textbf{ICF}                                & 1.30                          & 2.00                         & 0.84                          & 2.03    & 1.23                          & 2.01                        \\
BMS                                         & 1.29                          & 0.97                         & 0.87                          & 1.59    & 1.22                          & 1.07                        \\ \hline

\end{tabular}%
}
\caption{The top-3 classical and deep models (shown in bold) evaluated on the O\3 dataset.}
\label{tab:O3_results}
\end{table}

Due to space limitations, we report results only for the top-3 classical and deep models (Table \ref{tab:O3_results}) and present complete results in Section \ref{sec:o3_eval}. Given that the majority of targets in the O\3 dataset differ in several feature dimensions we do not list all possible combinations and discuss only subsets which do or do not differ in color since the presence of a color feature dominates other features during visual search \cite{kaptein1995search}.

\begin{figure*}[t]
\centering
\subfigure[]{
\includegraphics[height=0.32\textwidth]{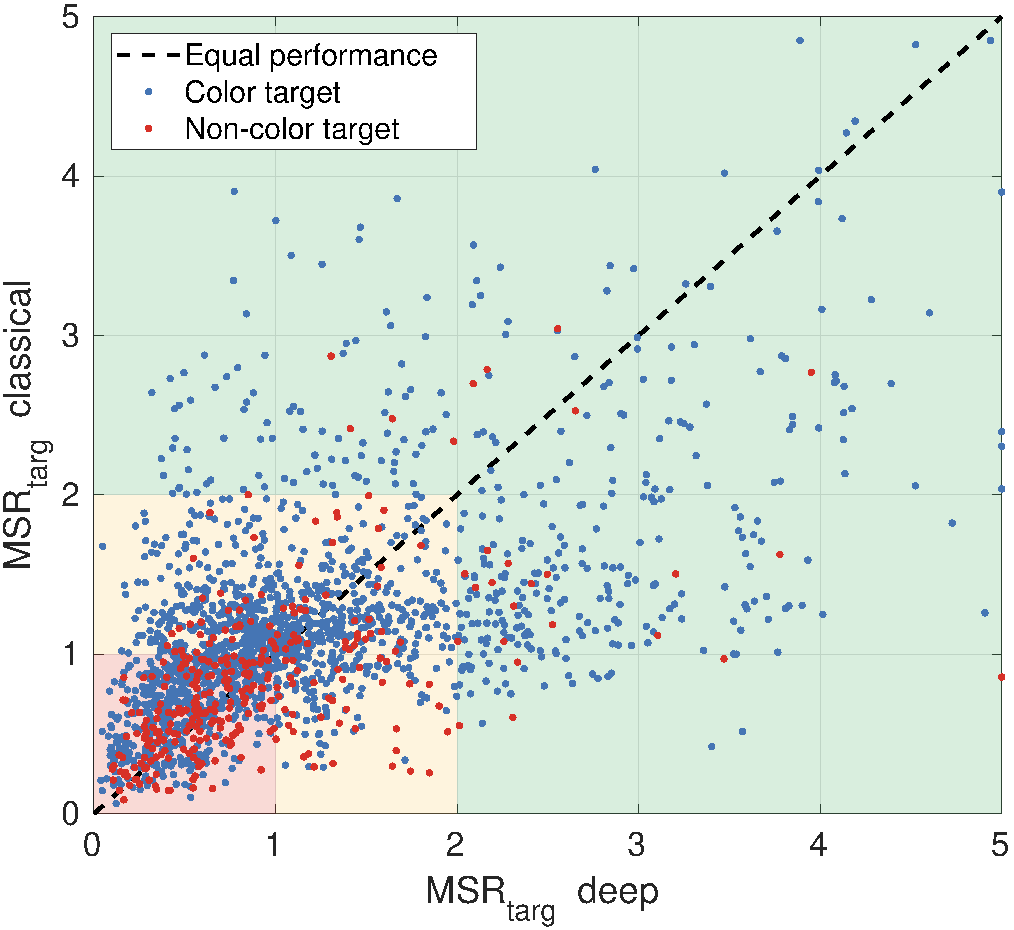}
\label{fig:O3_gsi}
}
\subfigure[]{
\includegraphics[height=0.32\textwidth]{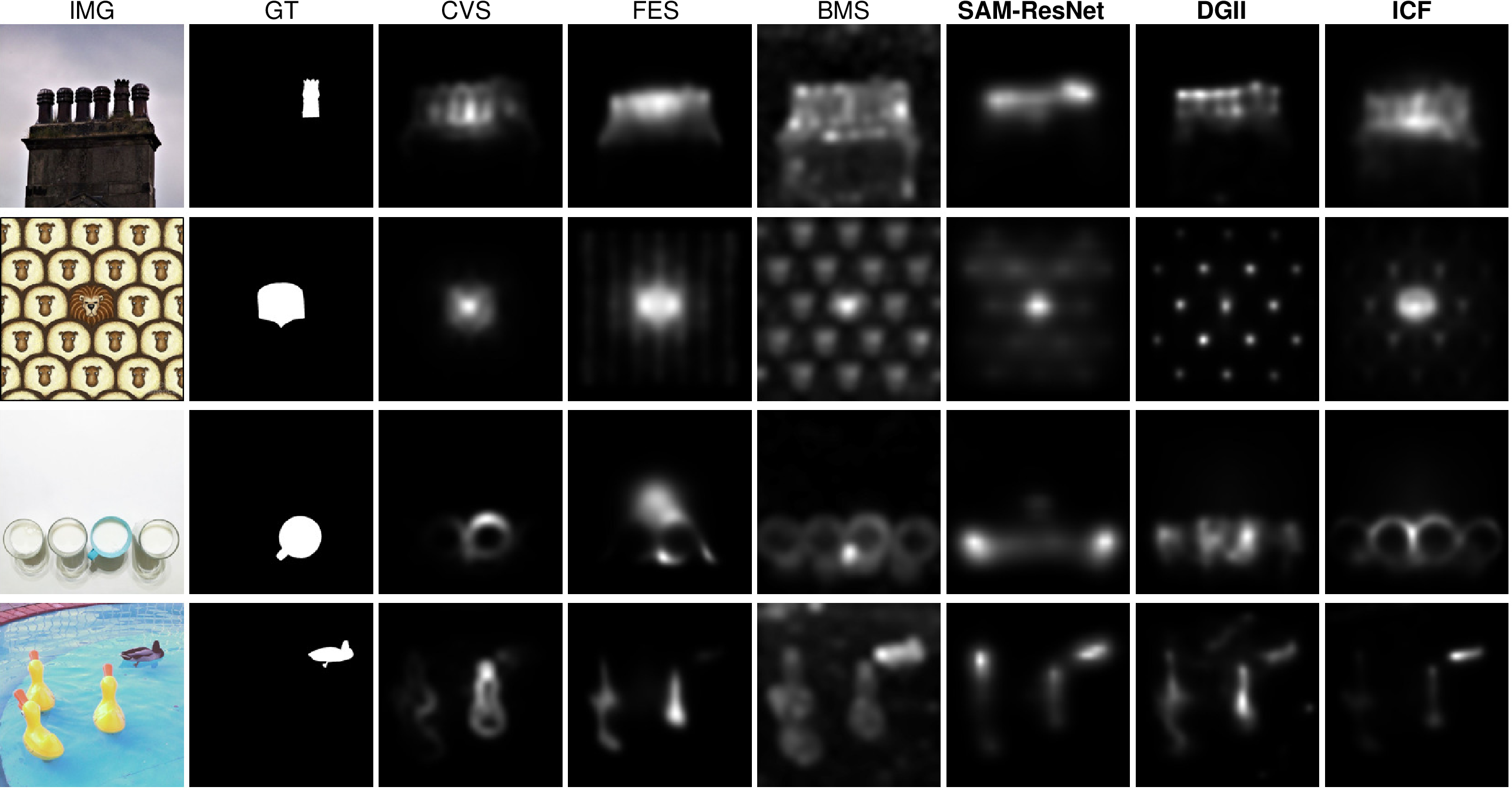}
\label{fig:O3_samples}
}

   \caption{a) The mean MSR$_{targ}$ of the top-3 classical and deep models for the color and non-color targets in O\3 (shown as blue and red dots respectively). Along the dashed line the performance of the models is equal. The red, yellow and green colors show areas where targets are not discriminated (MSR$_{targ} < 1$), somewhat discriminated ($1\leq$ MSR$_{targ}\leq 2$) and strongly discriminated (MSR$_{targ} > 2$). b) Sample images and corresponding saliency maps. From top to bottom row: hard for both, easy for both, classical models perform better, deep models perform better.}
\end{figure*}

All models discriminate color targets better than targets that differ in other feature dimensions, as indicated by low MSR$_{targ}$ values for non-color singletons in Table \ref{tab:O3_results} (see sample images and saliency maps in Figure \ref{fig:O3_samples}). As shown in Figure \ref{fig:O3_gsi} the majority (80\%) of non-color targets have an average  MSR$_{targ} < 1$ compared to only 30\% of color targets.

Classical and deep models discriminate 55\% and 45\% of all targets respectively. Deep models strongly discriminate only 15\% of the images (MSR$_{targ}$ is $>2$) compared to 10\% for classical models. However, for all models the average MSR$_{bg}$ scores are higher than the MSR$_{targ}$ scores (Table \ref{tab:O3_results}), i.e. saliency maxima in the background are higher than the maxima within the target mask meaning that these models get distracted by the background.

We also observe that most classical models perform better on P\3 with few exceptions such as CVS and IMSIG. For instance, BMS, which by design handles color features well, excells on P\3 and also benefits from an abundance of color targets in O\3. For deep learning models the opposite is true: models in general show much better results on O\3. In fact, SAM-ResNet which ranks first on O\3 (see Table \ref{tab:O3_results}) ranks second to last on P\3 (see Figure \ref{fig:found_vs_fixations}). One possible reason for such discrepancy is that SAM by design learns a strong center bias prior which works against it on P\3, where target distribution is uniform, but is an advantage on natural image datasets like O\3 which are usually more center-biased. Another consideration is that SAM and most other state-of-the-art saliency algorithms rely on features produced by a backbone CNN pre-trained on object classification task. While deep learning models capture high-level features more often associated with human gaze locations (e.g., faces, text), it may be the case that training data does not capture some of the fundamental properties exemplified in psychophysical stimuli. On the other hand, many classical models explicitly define salient regions as different from their surroundings based on a set of features and/or heuristics (e.g., AIM, IKN, GBVS, FES, CAS, CAS, SSR), and so are able to more efficiently detect target/distractor differences in the simple P\3 stimuli. This suggests that a combination of both theory- and data-driven approaches should be explored, similar to the ICF model that includes explicit local intensity and local contrast features as well as readout network trained on human gaze data. 

Overall, these results demonstrate that the inability of saliency models to adequately respond to singletons extends beyond synthetic stimuli and is also apparent in more realistic scenarios. This calls for changes in algorithm design and also has ramifications for the use of these computational models in practical applications mentioned in Section \ref{sec:intro}.

\subsubsection{Training Saliency Models on P\3 and O\3 Data}
As mentioned in Section \ref{sec:relevant_works}, fixation datasets contain a negligibly small portion of synthetic and real odd-one-out stimuli. Only CAT2000 has $\approx$4\% synthetic patterns but is rarely used for training. Specifically, only the SAM model is trained on CAT2000. Here, we conduct a simple test to determine whether the struggle of deep learning models to detect singletons in synthetic and real images can be corrected by augmenting the training data. We select MLNet and SALICON for further experimentation\footnote{MLNet, SAM and SALICON are the only models with publicly available training code, however SAM also requires fixation points for training which are not available for P\3 and O\3. We use our own implementation of SALICON (\url{https://github.com/ykotseruba/SALICONtf}) since open-source code \cite{thomas2016opensalicon} did not reproduce published results (see Section \ref{sec:salicon}).}.

\begin{table}[t]
\centering
\resizebox{1\textwidth}{!}{%
\begin{tabular}{l|l|lllll|ll|ll}
\multirow{2}{*}{Model}   & \multicolumn{1}{c|}{\multirow{2}{*}{Training Data}} & \multicolumn{5}{c|}{MIT1003}           & \multicolumn{2}{c|}{P\3} & \multicolumn{2}{c}{O\3} \\ \cline{3-11}
                         & \multicolumn{1}{c|}{}                               & AUC\_Judd & CC   & KLDiv & NSS  & SIM  & Avg. num. fix.           & \% found          & MSR$_{targ}$        & MSR$_{bg}$        \\ \hline
\multirow{6}{*}{MLNet}   & SALICON                                             & 0.82      & 0.47 & 1.3   & 1.7  & 0.35 & 40.97               & 0.46              & 0.92                 & 0.98             \\
                         & P\3                                  & 0.51      & 0.04 & 12.07 & 0.13 & 0.18 & 51.81               & 0.44              & 0.78                 & 2.77             \\
                         & O\3                                  & 0.74      & 0.29 & 1.83  & 1.01 & 0.3  & 41.32               & 0.46              & 1.01                 & 0.91             \\
                         & SALICON+P\3                          & 0.82      & 0.47 & 1.36  & 1.69 & 0.36 & 40.01               & 0.45              & 0.93                 & 0.97             \\
                         & SALICON+O\3                          & 0.82      & 0.46 & 1.33  & 1.65 & 0.34 & 40.87               & 0.46              & 0.97                 & 0.91             \\
                         & SALICON+P\3+O\3       & 0.82      & 0.46 & 1.36  & 1.64 & 0.35 & 42.00               & 0.44              & 0.96                 & 0.91             \\ \hline
\multirow{6}{*}{SALICON} & OSIE                                                & 0.86      & 0.6 & 0.92  & 2.12 & 0.48 & 57.12               & 0.68              & 0.88                 & 1.16             \\
                         & P\3                                  & 0.57      & 0.08 & 2     & 0.31 & 0.23 & 51.60               & 0.66              & 0.90                 & 1.32             \\
                         & O\3                                  & 0.77      & 0.32 & 1.52  & 1.08 & 0.34 & 45.37               & 0.70              & 1.05                 & 2.12             \\
                         & OSIE+P\3                             & 0.83      & 0.52 & 1.1   & 1.87 & 0.42 & 50.23               & 0.64              & 0.87                 & 1.27             \\
                         & OSIE+O\3                             & 0.85      & 0.58 & 1  & 1.98 & 0.45 & 45.90               & 0.72              & 0.91                 & 1.10             \\
                         & OSIE+P\3+O\3          & 0.83      & 0.51 & 1.12  & 1.84 & 0.41 & 49.37               & 0.65              & 0.90                 & 1.26
\end{tabular}%
}

\caption{Results of training MLNet and SALICON on different data.}
\label{tab:training_results}
\end{table}

We follow the original training procedures and only modify the training data. For MLNet we add 1553 P\3 and/or 1200 O\3 images to its training set of 10K images from the SALICON dataset. We augment the smaller OSIE dataset (700 images) used for training the SALICON model by adding 100 and 76 images from P\3 and O\3 respectively, thus matching the proportion of extra P\3 and O\3 data of 16\% and 12\%, respectively, for both models. We also train both models only on P\3 and O\3 to test whether the task can be learned from these examples. We use target binary masks as a substitute for human fixation maps. We evaluate models on 518 P\3 images and 401 O\3 images not used for training and on the MIT1003 dataset.

Table \ref{tab:training_results} summarizes the results of training MLNet and SALICON on various combinations of the original training data and data from P\3 and O\3. According to the evaluation against human fixations for the MIT1003 dataset (which is a reasonable approximation of performance on the MIT300 benchmark), augmenting training data with P\3 and/or O\3 patterns does not have a significant effect on fixation prediction accuracy. Both models generally do not learn well from the P\3 data alone and results are the worst overall. The O\3 training data results in the best performance on the O\3 test data, which is expected, but also leads to relatively high performance on the MIT1003 data as well as P\3 patterns. One possible explanation for the poor performance from P\3 compared to O\3 is that synthetic stimuli considerably differ in appearance from natural images making it difficult for CNN-based models to effectively leverage object representaions learnt in object classification (\eg VGG\cite{simonyan2014very}).

The best results are achieved when O\3 is mixed with the original training set for both models. Note that training on the O\3 data leads to the highest MSR$_{targ}$ and MSR$_{targ}$ > MSR$_{bg}$ for both MLNet and SALICON models, meaning that they somewhat improved their ability to distinguish targets and are less distracted by the objects in the background.

Our experiment shows that only minor improvements can be achieved by augmenting the training datasets with odd-one-out data and suggests that other factors might be at play. This is in agreement with literature exploring the behavior of deep-learning models in a related category of tasks requiring comparisons between items (\eg the \textit{same-different} task to determine whether the presented objects are the same or not). These works explore a range of architectures and determine that even training on datasets several orders of magnitude larger than ours does not bring these models close to human-level \cite{stabinger201625, gulccehre2016knowledge, kim2018not}. In particular, in \cite{kim2018not} the authors hypothesize that feedforward architectures lack mechanisms for binding features to individuated objects which makes them struggle with these types of tasks. However, further research is needed to determine whether these conclusions can be extended to the odd-one-out detection task.

\section{Conclusion}

Despite saturated performance on common benchmarks, most classical and deep models still do not adequately model fundamental attention guiding mechanisms. Using two newly proposed datasets, P\3 and O\3, we systematically examined the performance of the saliency algorithms on synthetic stimuli similar to the ones used in human research and on a large collection of realistic odd-one-out images. We showed that several classical models discriminate targets in some of the feature dimensions considered and also perform better on average than the CNN-based models. However, a large gap remains between human and algorithm performance both quantitatively and qualitatively. We also showed that augmenting training data for two CNN-based saliency models or training them exclusively on P\3 and O\3 images does not significantly improve their ability to discriminate singletons and has only a minor positive effect on fixation prediction accuracy, which requires further investigation.

\section*{Acknowledgements}
This work was supported by the Natural Sciences and Engineering
Research Council of Canada (NSERC), the NSERC Canadian Robotics Network (NCRN), the Air Force Office for Scientific Research (USA), and the Canada Research Chairs Program through grants to John K. Tsotsos.



\section{Supplementary Materials}

\subsection{Number of Psychophysical Patterns and Natural Odd-One-Out Images in Common Fixation Datasets}
\label{sec:data_stats}

Table \ref{tab:sample_counts} lists 10 datasets with human fixation data commonly used for training and evaluating computational saliency models. We examined these datasets and recorded for each the number of synthetic psychophysical patterns and natural odd-one-out images they contain. 

\begin{table}[th]
\centering
\begin{tabular}{l|c|c|c}
Dataset   & \multicolumn{1}{l|}{\begin{tabular}[c]{@{}l@{}}Num. of synthetic \\ psychophysical \\ patterns\end{tabular}} & \multicolumn{1}{l|}{\begin{tabular}[c]{@{}l@{}}Num. of natural \\ odd-one-out images\end{tabular}} & \multicolumn{1}{l}{\begin{tabular}[c]{@{}l@{}}Total \\ number \\ of images\end{tabular}} \\ \hline
SID4VAM \cite{berga2019psychophysical}   & 230                                                                                         & 0                                                                                                  & 230                                                                                     \\
CAT2000 \cite{CAT2000}   & 71                                                                                         & 7                                                                                                  & 2000                                                                                     \\
Toronto \cite{bruce2007attention}   & 16                                                                                         & 4                                                                                                  & 136                                                                                      \\
MIT1003 \cite{judd2009learning}  & 4                                                                                          & 6                                                                                                  & 1003                                                                                     \\
SALICON  \cite{jiang2015salicon} & 0                                                                                          & 33                                                                                                 & 20000                                                                                    \\
DUT-OMRON \cite{yang2013saliency} & 0                                                                                          & 19                                                                                                 & 5168                                                                                     \\
NUSEF  \cite{ramanathan2010eye}   & 0                                                                                          & 3                                                                                                  & 447                                                                                      \\
VIU    \cite{koehler2014saliency}   & 0                                                                                          & 2                                                                                                  & 800                                                                                      \\
OSIE   \cite{xu2014predicting}   & 0                                                                                          & 1                                                                                                  & 700                                                                                      \\
Ehinger \cite{ehinger2009modelling}  & 0                                                                                          & 0                                                                                                  & 912                                                                                      \\ \hline
\end{tabular}
\caption{Number of synthetic patterns and natural odd-one-out images in the commonly used fixation datasets and benchmarks.}
\label{tab:sample_counts}
\end{table}

\subsection{Generating Synthetic Search Arrays for P\3 dataset}
\label{sec:data_gen}

Below are detailed descriptions of images generated for each feature dimension in the P\3 dataset.

\textbf{Orientation (864 images)}. The following target/distractor shapes are used: rectangular bar, ellipse, tree and map-marker. The angle of the target is varied from 0\degree to 180\degree with a step of 15\degree. For each target orientation the distractors are rotated so that the target/distractor orientation difference varies from -90\degree to 90\degree with a step of 10\degree (omitting 0\degree difference). The total number of images is [$4$ shapes] $\times$ [$11$ target orientations] $\times$ [$18$ target/distractor orientation differences] = 864. Stimuli are rendered in dark navy (RGB(2,56,88)) on a white background.

\textbf{Size (840 images)}. The size of the target is varied between 0.25 to 2$\times$ the size of the distractors (75 px) with a step of 0.25. Both target and distractors are rotated from 0\degree to 180\degree with a step of 6\degree. Background and stimuli colors are the same as in the orientation set. We use square, circle, rectangular bar and ellipse shapes. The total number of images in this set is [$4$ shapes] $\times$ [$30$ orientations] $\times$ [$7$ target/distractor size ratios] = 840.

\textbf{Color (885 images)}. Following the procedure proposed in \cite{wloka2017interaction}, 36 colors are selected on a circle of radius 60 around the point (58, 12, 13) in the CIELAB space. The target and distractors colors are set so that the difference between them varies between -180\degree and 180\degree with a step of 6\degree (omitting 0\degree difference). Shapes of the elements are set to rectangular bar, square or circle. We also rotate both targets and distractors from -90\degree to -90\degree with a step of 45\degree. Overall there are [$3$ shapes] $\times$ [$5$ rotations] $\times$ [$59$ target/distractor color differences] = 885 color images. Background is set to grey (RGB(128, 128, 128)).

In all images the target location is selected at random.

\subsection{Evaluation on Psychophysical Stimuli in P\3 Dataset}
\label{sec:psych_eval}

\begin{figure*}[th]
\centering
\subfigure[color difference]{
\includegraphics[width=0.3\textwidth]{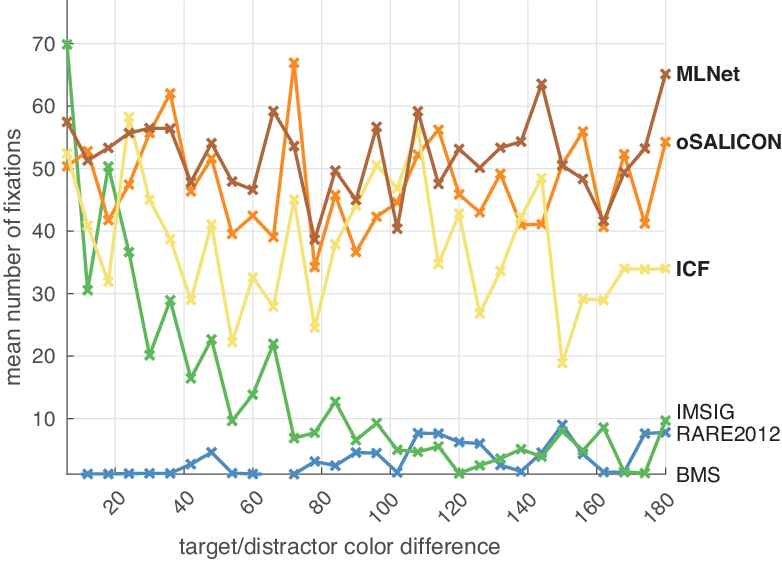}
\label{fig:abs_hue_angle_diff}
}
\subfigure[orientation difference]{
\includegraphics[width=0.3\textwidth]{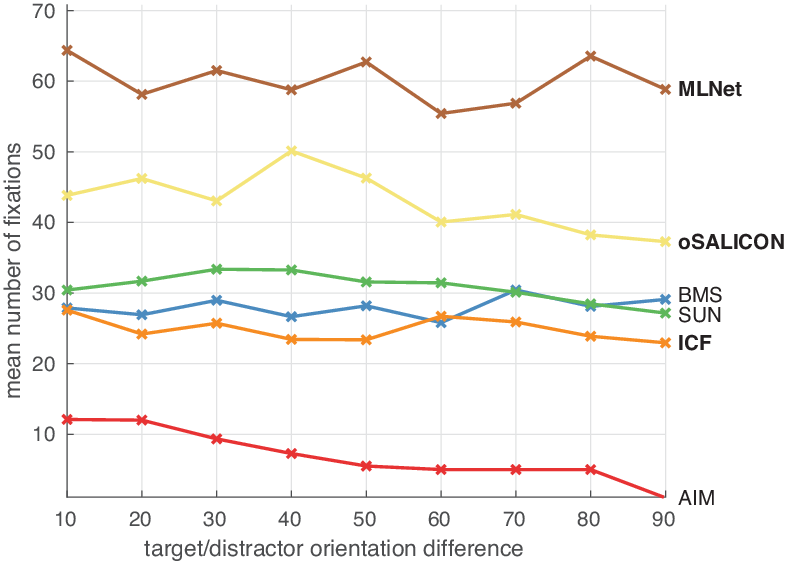}
\label{fig:abs_orientation_diff}

}
\subfigure[size ratio]{
\includegraphics[width=0.3\textwidth]{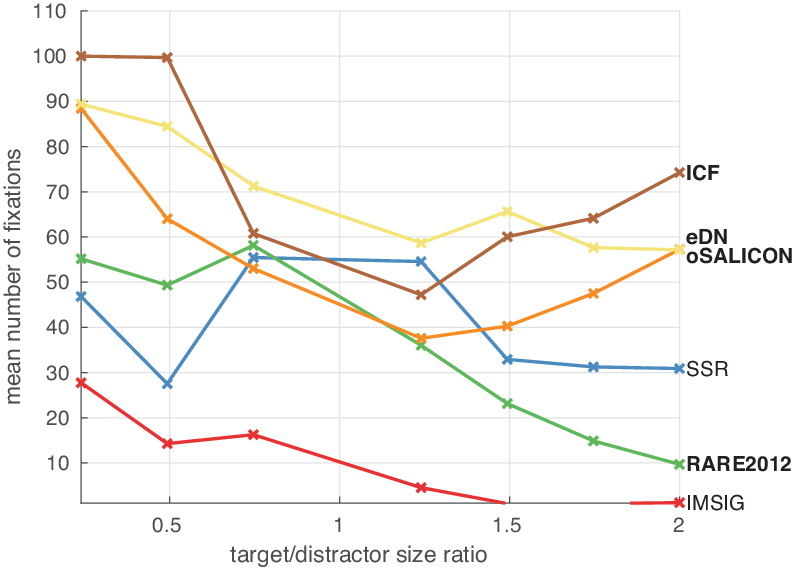}
\label{fig:size_ratio}
}
   \caption{Average number of fixations-to-target for the top-3 classical and deep models for a range of TD differences in color (a), orientation (b) and size (c) feature dimensions. The top models are selected based on the total number of targets found within 100 fixations in each dimension. }
\label{fig:num_fixations}
\end{figure*}

\begin{figure*}[t!]
\centering
\subfigure[color difference]{
\includegraphics[height=0.4\textheight]{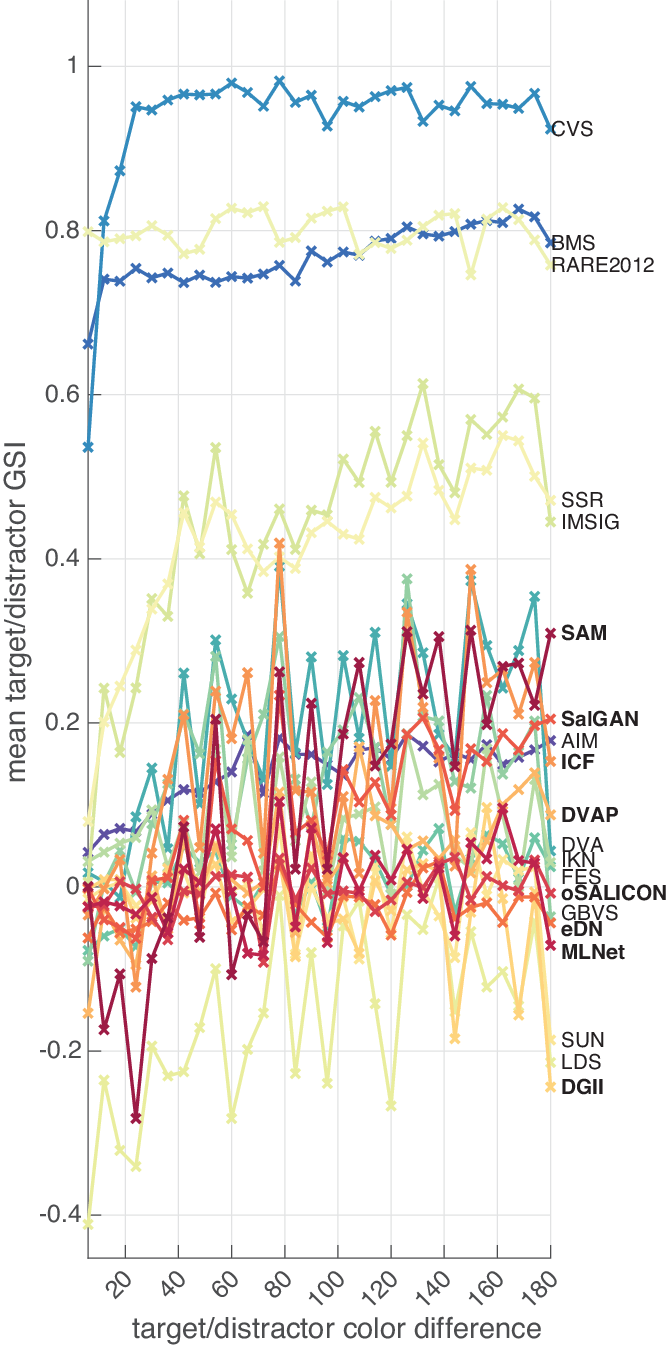}
\label{fig:all_abs_hue_angle_diff}
}
\subfigure[orientation difference]{
\includegraphics[height=0.4\textheight]{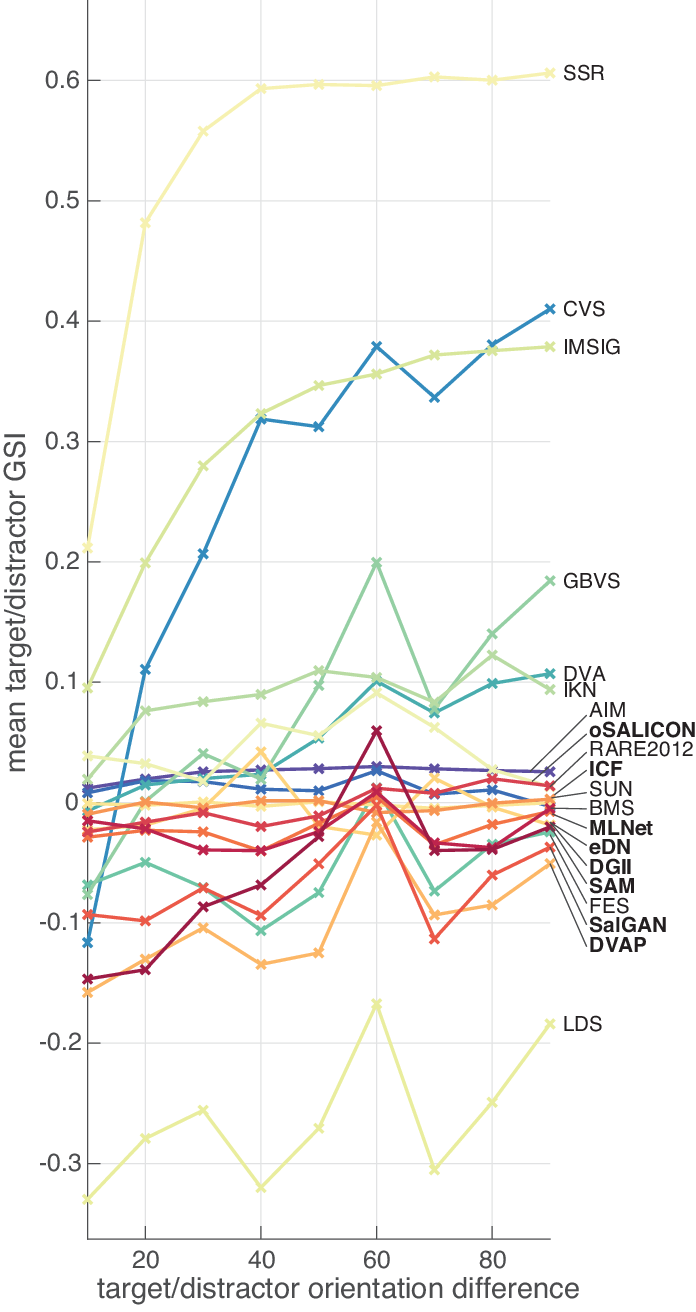}
\label{fig:all_abs_orientation_diff}

}
\subfigure[size ratio]{
\includegraphics[height=0.4\textheight]{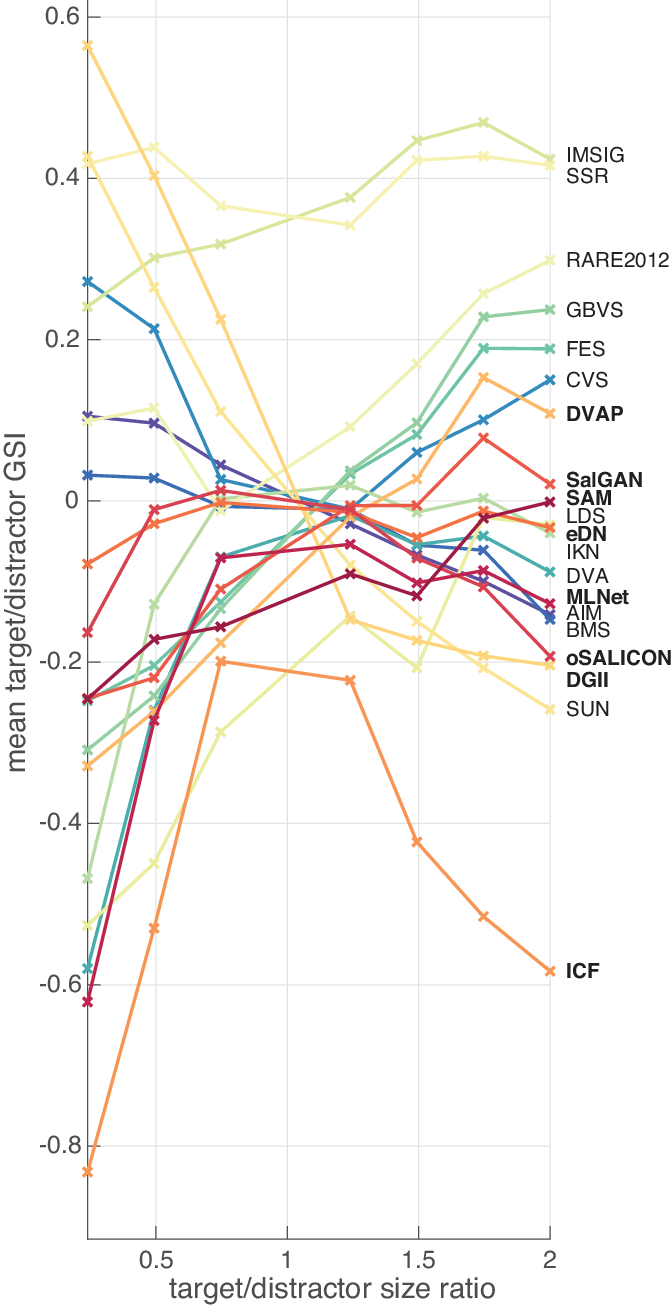}
\label{fig:all_abs_size_ratio}
}
   \caption{Plots of discriminative ability of 12 classical and 8 deep models for a range of TD differences in color (a), orientation (b) and size (c) feature dimensions according to the GSI metric. Deep-learning model labels are rendered in bold.}
\label{fig:all_search_feature}
\end{figure*}

Figure \ref{fig:num_fixations} shows the relationship between the target-distractor difference and number of fixations needed to reach the target for top-3 classical and top-3 deep-learning algorithms to support our argument in Section 4.1.1.

Figure \ref{fig:all_search_feature} shows the discriminative ability of 12 classical and 8 deep models for different features according to GSI metric. Orientation is particularly challenging for many models as many of them cluster between 0.1 and 0 GSI (Figure \ref{fig:abs_orientation_diff}).

Note that even though some algorithms show good discrimination ability on some features (\eg CVS and SSR) they also miss a significant portion of the targets and therefore are not featured as top-3 performing models discussed in Section 4.1.1.

\begin{figure*}[t!]
\centering
\subfigure{
\includegraphics[height=0.75\textheight]{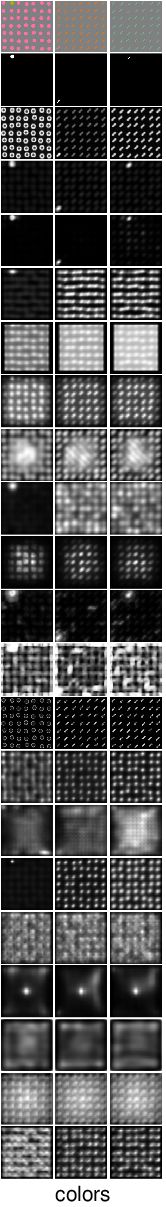}
\label{fig:P3_sample_images_colors}
}
\subfigure{
\includegraphics[height=0.75\textheight]{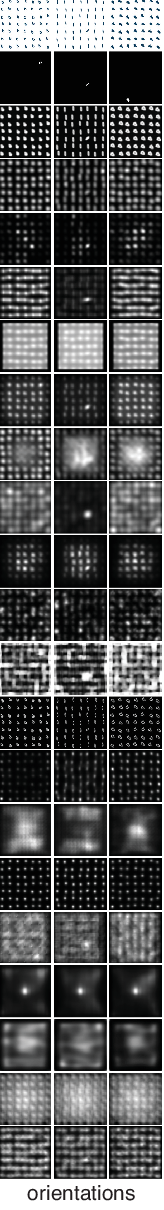}
\label{fig:P3_sample_images_orientations}

}
\subfigure{
\includegraphics[height=0.75\textheight]{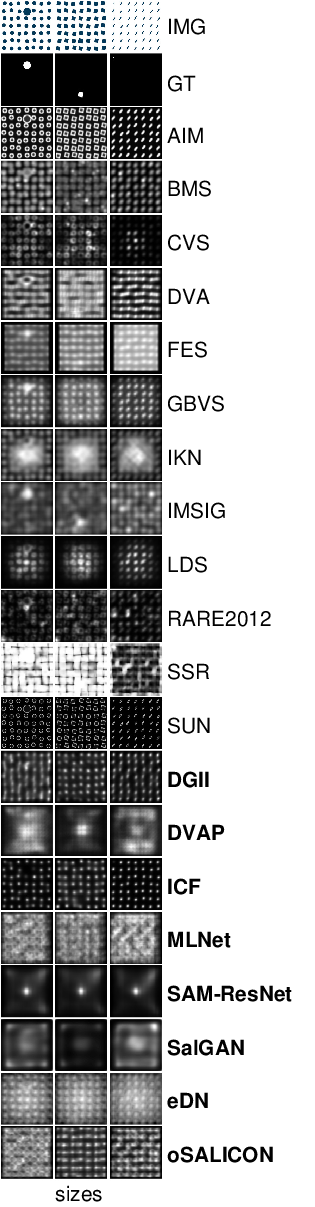}
\label{fig:P3_sample_images_sizes}
}
   \caption{Examples of saliency maps for 12 classical and 8 deep models for a range of TD differences in color (a), orientation (b) and size (c) feature dimensions. Models are sorted alphabetically. Deep-learning model labels are rendered in bold.}
\label{fig:P3_saliency_maps}
\end{figure*}

Figure \ref{fig:P3_saliency_maps} shows sample saliency maps produced by classical and deep-learning algorithms for color, orientation and size. For each feature we selected examples representing a range of target-orientation differences. 

\clearpage
\subsection{Evaluation on Odd-One-Out Stimuli in O\3 dataset}
\label{sec:o3_eval}

Table \ref{tab:o3_results} shows evaluation results for 12 classical and 8 deep-learning models on the O\3 dataset. We report the performance for color and non-color targets separately, as well as the average performance for all targets. Note that for nearly all models average MSR$_{bg}$ scores are higher than MSR$_{targ}$ (sometimes significantly) meaning that the maximum saliency values in the background are higher than those of targets.

\begin{figure*}[th]
\centering

\includegraphics[height=0.93\textheight]{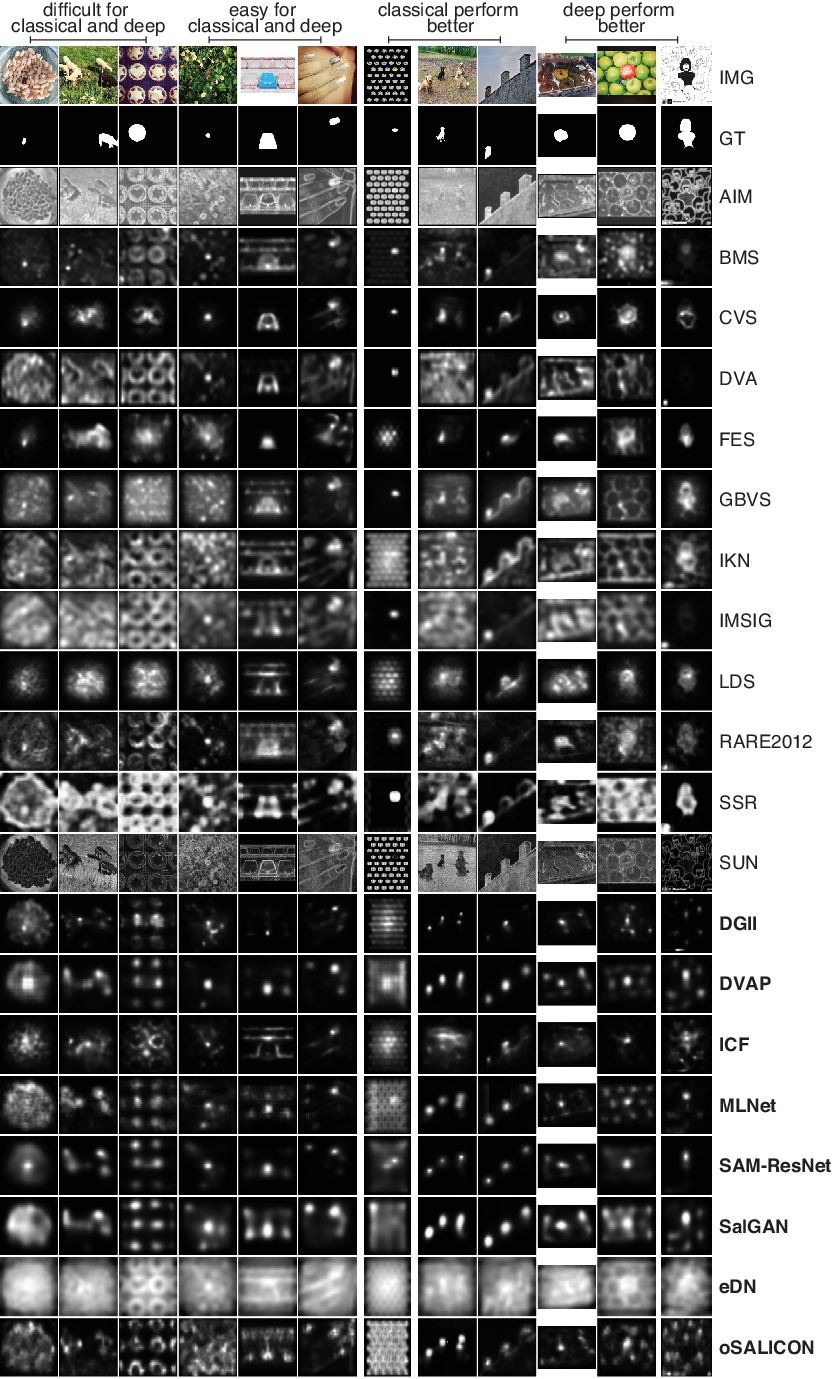}
\caption{Sample images and corresponding saliency maps. 3 samples are selected for the following target categories: easy for classical and deep models, difficult for both, classical models perform better, deep models perform better. Classical and deep models are sorted alphabetically. Deep model labels are rendered in bold.}
\label{fig:O3_samples_all}
\end{figure*}

Figure \ref{fig:O3_samples_all} shows sample images and saliency maps for a range of natural odd-one-out targets. We select 3 examples in each category based on average MSR$_{targ}$: difficult for classical and deep models (MSR$_{targ}<1$), easy for classical and deep models (MSR$_{targ}>2$), easier for classical models (MSR$_{targ}$ for classical models is > 1 but not for deep ones) and easier for deep models (MSR$_{targ}$ for deep models is > 1 but not for classical ones).

\begin{table}[th]
\resizebox{\textwidth}{!}{%
\begin{tabular}{l|ll|ll|ll}
\multicolumn{1}{c|}{\multirow{2}{*}{Model}} & \multicolumn{2}{c|}{Color targets} & \multicolumn{2}{c|}{Non-color targets} & \multicolumn{2}{c}{All targets} \\ \cline{2-7} 
\multicolumn{1}{c|}{}                       & MSR$_{targ}$     & MSR$_{bg}$ & MSR$_{targ}$       & MSR$_{bg}$      & MSR$_{targ}$    & MSR$_{bg}$   \\ \hline
\textbf{SAM-ResNet}                         & 1.47              & 1.46           & 1.04                & 1.84             & 1.40             & 1.52          \\
CVS                                         & 1.43              & 2.43           & 0.91                & 4.26             & 1.34             & 2.72          \\
\textbf{DGII}                               & 1.32              & 1.55           & 0.94                & 1.95             & 1.26             & 1.62          \\
FES                                         & 1.34              & 2.53           & 0.81                & 5.93             & 1.26             & 3.08          \\
\textbf{ICF}                                & 1.30              & 2.00           & 0.84                & 2.03             & 1.23             & 2.01          \\
BMS                                         & 1.29              & 0.97           & 0.87                & 1.59             & 1.22             & 1.07          \\
RARE2012                                    & 1.18              & 0.98           & 0.83                & 1.54             & 1.13             & 1.07          \\
DVA                                         & 1.13              & 1.28           & 0.81                & 1.70             & 1.08             & 1.35          \\
GBVS                                        & 1.07              & 1.01           & 0.89                & 1.29             & 1.04             & 1.06          \\
IMSIG                                       & 1.05              & 1.10           & 0.91                & 1.54             & 1.03             & 1.17          \\
SSR                                         & 1.05              & 1.09           & 0.91                & 2.31             & 1.03             & 1.29          \\
\textbf{DVAP}                               & 1.05              & 1.22           & 0.86                & 1.41             & 1.02             & 1.25          \\
\textbf{MLNet}                              & 1.00              & 1.03           & 0.86                & 1.33             & 0.98             & 1.08          \\
LDS                                         & 1.01              & 2.02           & 0.76                & 2.17             & 0.97             & 2.05          \\
AIM                                         & 0.97              & 0.94           & 0.96                & 1.03             & 0.97             & 0.95          \\
\textbf{eDN}                                & 0.95              & 0.94           & 0.92                & 1.05             & 0.94             & 0.96          \\
SUN                                         & 0.93              & 1.03           & 0.91                & 1.13             & 0.93             & 1.05          \\
\textbf{SalGAN}                             & 0.92              & 1.22           & 0.89                & 1.41             & 0.92             & 1.25          \\
\textbf{oSALICON}                           & 0.89              & 1.19           & 0.83                & 1.40             & 0.88             & 1.23          \\
IKN                                         & 0.88              & 1.16           & 0.83                & 1.39             & 0.87             & 1.20          \\ \hline
\end{tabular}%
}
\label{tab:o3_results}
\caption{Evaluation results for 12 classical and 8 deep-learning saliency algorithms on the O\3 dataset. The models are sorted according to average MSR$_{targ}$ measure. Deep-learning model names are rendered in bold.} 
\end{table}

\clearpage
\subsection{SALICON Implementation}
\label{sec:salicon}
\begin{table}[t]
\centering
\resizebox{0.8\textwidth}{!}{%
\begin{tabular}{l|l|l|l|l|l}
\multicolumn{1}{c|}{\multirow{2}{*}{Model}} & \multicolumn{5}{c}{MIT1003 Dataset}   \\ \cline{2-6} 
\multicolumn{1}{c|}{}                       & AUC\_Judd & CC   & KLDiv & NSS  & SIM  \\ \hline
SALICON (online demo)                       & 0.87      & 0.62 & 0.96  & 2.17 & 0.5  \\
OpenSALICON (published)                     & 0.83      & 0.51 & 1.14  & 1.92 & 0.41 \\
OpenSALICON (trained)                       & 0.51      & 0.03 & 2.06  & 0.13 & 0.22 \\
SALICONtf (ours)                            & 0.86      & 0.6  & 0.92  & 2.12 & 0.48 \\ \hline
\end{tabular}%
}
\label{tab:salicon_results}
\caption{Evaluation results for different versions of SALICON model on the MIT1003 dataset. Online demo: saliency maps were obtained from the official SALICON online demo at \url{http://salicon.net/demo/}. Published: saliency maps obtained using the pre-trained OpenSALICON weights (\url{https://github.com/CLT29/OpenSALICON}). Trained: saliency maps produced by the model trained using the code provided with the OpenSALICON weights. Ours: our reimplementation of the SALICON model in TensorFlow (\url{https://github.com/ykotseruba/SALICONtf}).}
\end{table}

We reimplement the SALICON model since the training code provided with the open-source version, OpenSALICON \cite{thomas2016opensalicon}, did not reach the performance of the published model as shown in Table  \ref{tab:salicon_results}. We use MIT1003 dataset for evaluation since it best reflects the performance on the official MIT300 benchmark \cite{mit-saliency-benchmark}.

\textbf{Architecture.} As in the original formulation of the model \cite{huang2015salicon} we use a two-stream VGG network \cite{simonyan2014very} without top fully-connected layers.  Our implementation uses TensorFlow \cite{abadi2016tensorflow} instead of caffe \cite{jia2014caffe}, hence the name SALICONtf. The coarse stream receives inputs resized to 300$\times$400 px and the fine stream receives inputs resized to 600$\times$800 px. (Note that in the OpenSALICON model the input sizes are set to 600$\times$800 px and 1200$\times$1600 px for the coarse and fine streams respectively.) The final layer of the fine stream is resized to match the size of the coarse stream (37$\times$50 px). Both outputs are concatenated and convolved with 1$\times$1 filter. The labels (human fixation maps) are resized to 37$\times$50 to match the output of the network.

\textbf{Training.} In the original formulation the best results were achieved by optimizing the Kullback-Leibler divergence (KLD) loss. In our experiments with SALICONtf we obtained better results using the binary cross-entropy loss as in \cite{thomas2016opensalicon}. We use fixed learning rate of $10^{-3}$, momentum of 0.9 and weight decay of 0.0005. Note that the original paper did not specify the number of training epochs and only mentioned that between 1 and 2 hours is required to train the model. Our implementation achieves reasonable results after 100 epochs and reaches its top perfomance on MIT1003 dataset at 300 epochs.

The model is trained on the OSIE dataset, which we split into training set of 630 images and validation set of 70 images. We evaluate the model on MIT1003 dataset. The results in Table \ref{tab:salicon_results} show that our model achieves results closest to the official SALICON demo results. The model runs at $\approx 5$ FPS on the NVIDIA Titan X GPU.

\bibliography{psychosal}
\end{document}